\documentclass[letterpaper, 10 pt, conference]{ieeeconf}  

\usepackage{xcolor}
\usepackage{dsfont}
\usepackage{amsmath,amssymb,amsfonts}
\usepackage{algorithmic}
\usepackage{multirow}
\usepackage{graphicx,graphics}
\usepackage{caption,subcaption}
\usepackage{xcolor}
\usepackage{url}
\usepackage{booktabs}
\usepackage{tabularx}
\usepackage{bm}
\usepackage{siunitx}

\IEEEoverridecommandlockouts                              

\title{\LARGE \bf
Predicting Maximum Permitted Process Forces for Object Grasping and Manipulation Using a Deep Learning Regression Model
}

\author{Stefanie Wucherer$^{1}$, Robert McMurray$^{2}$, Kok Yew Ng$^{2,3}$, Florian Kerber$^{1}$
	\thanks{$^{1}$S. Wucherer and F. Kerber are with the Technology Transfer Center of the Technical University of Applied Sciences Augsburg, Germany.
		Emails: {\tt \footnotesize \{stefanie.wucherer, florian.kerber\}@hs-augsburg.de}.}%
	\thanks{$^{2}$K. Y. Ng and R. McMurray are with the Engineering Research Institute of Ulster University, UK.
		Emails: {\tt \footnotesize \{mark.ng, rj.mcmurray\}@ulster.ac.uk}.}%
	\thanks{$^{3}$K. Y. Ng is also affiliated with the School of Engineering, Monash University, 47500 Selangor, Malaysia. 
		Email: {\tt \footnotesize kok.yew.ng@monash.edu}.}
}

\begin{document}
	\maketitle
	\thispagestyle{empty}
	\pagestyle{empty}
	
	\begin{abstract}
		During the execution of handling processes in manufacturing, it is difficult to measure the process forces with state-of-the-art gripper systems since they usually lack integrated sensors. Thus, the exact state of the gripped object and the actuating process forces during manipulation and handling are unknown. 
        This paper proposes a deep learning regression model to construct a continuous stability metric to predict the maximum process forces on the gripped objects using high-resolution optical tactile sensors. 
        A pull experiment was developed to obtain a valid dataset for training. 
        Continuously force-based labeled pairs of tactile images for varying grip positions of industrial gearbox parts were acquired to train a novel neural network inspired by encoder-decoder architectures. A ResNet-18 model was used for comparison. 
		Both models can predict the maximum process force for each object with a precision of less than \SI{1}{N}.
		During validation, the generalization potential of the proposed methodology with respect to previously unknown objects was demonstrated with an accuracy of  \numrange[range-phrase = --]{0.4}{2.1}~\SI{}{N} and precision of \numrange[range-phrase=--]{1.7}{3.4}~\SI{}{N}, respectively. 
	
	\end{abstract}

	\section{Introduction}\label{sec:Intro}
	Deriving the stability of grip has been an active research area for over 35 years \cite{jameson1986}. 
	In initial approaches, it is assumed that the object geometry is known and the point of contact is given \cite{montana1991condition}. 
	To robustly automate industrial handling and assembly tasks, it is crucial to accurately predict the quality of a grip in the subsequent process. This prediction helps to determine whether to proceed with the task or to regrip, as failed attempts may result in time and monetary losses, disruptions to the assembly line, or cause damage to parts or the robot. Contemporary robots lack the necessary sensor feedback for this prediction, as they only receive coarse force feedback information. 
	
	Some existing papers investigated methods to derive strategies to evaluate if a performed grip is stable on an unknown object. For example, Hogan et al. \cite{hogan2018tactileRegrasp} presented a stability metric using optical tactile sensors where unknown objects are lifted with a grip force of \SI{30}{N} for undefined grip points. The images acquired are labeled via a shaking experiment. 
	A normalized stability measure is then applied to the tactile images using the measured and maximum time. The metric is obtained by training a ResNet-50 network \cite{he2016deep} using the labeled data. 
	Meanwhile, Calandra et al. \cite{calandra2018gelsight} introduced a gripping strategy where a multimodal residual network was trained to predict the grip success probability after an adjustment action candidate has been performed by using the tactile and visual information from the GelSight sensors and an RGB camera during an executed grip. 
	In Si et al. \cite{si2022taxim}, two approaches were used to predict stability: single-image and multi-image methods; both cases use binary labels to distinct between stable and non-stable grips. For the single-image method, a pre-trained ResNet-50 \cite{he2016deep} network is used whereas for the multi-image method, the images are passed to a long short-term memory (LSTM) module \cite{greff2017LSTM} where the output is then fed to a classifier. In their experiment, 12 household objects with different masses were subjected to \numrange[range-phrase = --]{5}{10}~\SI{}{N} of grip force and lifted using an UR5 robot with a WSG-50 gripper installed (with one GelSight and a 3D-printed finger) at different positions. During the process, the algorithm would check if the object has been successfully lifted. 
	A digital twin was also built for this setup to model physical properties such as the friction coefficient between object and gripper, as well as the perceptual properties of the GelSight Tactile Finger. Data generated from the twin was then used to train the neural network (NN) models.
	Then, Kolamuri et al. \cite{kolamuri2021} utilized an analytical approach to detect the rotation of a gripped object while it is being lifted using GelSight sensors with matrices of dots on the gel surface. 
	From the displacement of the dots and the calculated rotation, the torque and therefore the displacement of the grip from the center of mass could be estimated, allowing a regripping action to safely grip the object without rotation. 
	In 2023, Zhang et al. \cite{zhang2023_graspstability} utilized an attention guided, cross-modality fusion architecture by fusing visual and tactile sensor information to derive the stability of random grip points on household objects. The data sample was acquired via simulation where the arbitrary grasp configurations varied in grip force between $5$ and \SI{35}{N}. After a successful grasp, the object was shaken by the robot and a binary label was assigned by determining if the object was still held.

	In this paper, we introduce a continuous force-based metric using RGB images from GelSight tactile sensors to predict the maximum force applicable during task execution. 
	To achieve that, we developed an experiment to obtain a reliable dataset, mimicking conditions especially static loads during assembly processes. 
	Industrial gearbox parts are used to generate the grip prediction dataset, instead of household items that are commonly used as benchmarks in the literature.
	The novel approach of using continuous labels, along with a wide variety of grip points and forces, refine the prediction from a mere success probability to an actual measurement of permitted process force application. This allows for more granular decision-making and even process control.
	
	
	This paper is organized as follows: Section \ref{sec:ProbState} introduces the problem statement; Section \ref{sec:methodology} presents the general overview of the proposed methodology; Section \ref{sec:experiment} presents the automated pull force experimental setup; Section \ref{sec:network_design} demonstrates the design and training of the NN; Section \ref{sec:validation} discusses results from the validation of the trained models; and Section \ref{sec:conclusion} provides some conclusions.

	\section{Problem Statement}\label{sec:ProbState}
	We want to predict force limits for the manipulation of gripped objects during assembly, in particular, joining tasks like peg-in-hole processes. 

	Specialized sensors have since been developed to reliably measure the state of the object that allow for refined predictions. One such sensor is the GelSight R1.5 \cite{yuan2017}, a tactile sensor that is able to measure the contact topology, which can then be correlated with the maximum process force permitted. 
	
	Robotic assembly tasks such as peg-in-hole processes require object manipulation and joining of objects. Gripped objects thus experience both dynamic forces during manipulation and joining as well as static forces, in particular due to gravity. 
	Our main goal is to develop a stability metric to predict the maximum permitted process forces for gripped objects during manipulation, such that the object does not slip. 

	Only industrial assembly parts are considered in this paper, however the methodology is not restricted to any domain. 
	Given that contact dynamics are often highly nonlinear and object characteristics such as geometry and material properties can vary, a deep learning approach is adopted. 
	To generate the necessary training data, an experimental setup is designed to measure maximum process forces of gripped objects. 
	In this paper, the experiment is set up such that the point of attack of the process forces and the object position are aligned, see Figure \ref{fig:setup}, such that no torques arise. 
	It should be emphasized that more general setups for complex contact or manipulation scenarios can be treated analogously. 
	The trained networks are validated using previously unknown objects. 
	
	\section[Stability Metric: Methodology]{Methodology}\label{sec:methodology}
	During manipulation and joining, an object endures process forces and torques represented as a wrench $\mathbf{w}_{\rm ext}=[F_x,F_y,F_z,M_x,M_y,M_z]$ where $F_i$ with $i\in\{x,y,z\}$ denotes the forces along the x-, y-, and z-directions while $M_i$ denote the torques around the respective axes. 
	A general stable grip can be defined as a set of wrenches $W_{\rm grip}=\{\mathbf{w}_1,...,\mathbf{w}_n\}$ applied to a set of points of attack $X_{\rm{grip}}=\{\bm{\xi}_1,...,\bm{\xi}_n\}$, where $\bm{\xi}_i=[\mathbf{x}_i,\mathbf{p}_i]\in SE(3)$, one for each of the $n$ fingers of the gripper, called the grip configuration $C=\{(\bm{\xi}_1,\mathbf{w}_1),...,(\bm{\xi}_n,\mathbf{w}_n)\}$. 

	Slippage occurs if the object moves within the jaws. More formally, we define a stability region $U_{r,\alpha}(\bm{\xi}_i)\in SE(3)$. For the grip point $\mathbf{x}_i\in \mathbb{R}^3$, this constitutes the hull within radius $r$ around $\mathbf{x}_i$. For the rotation $\mathbf{p}_i \in SO(3)$ this is the subset of all rotations $\mathbf{q}\in SO(3)$ fulfilling the condition $d_g(\mathbf{q},\mathbf{p}_i)<\alpha$ where $d_g$ is the geodesian distance and $\alpha$ an angle.
	We want to predict the maximum process forces and torques of the external wrench $\mathbf{w}_{\rm ext}$ for which the gripped object remains within the stability region $U_{r,\alpha}(\bm{\xi}_i)\ \forall i$ for an explicit grip configuration $C$, i.e., slippage does not occur.
	This prediction should be made from the grip topology $I$ obtained via the RGB images of the GelSight R1.5 tactile sensors at configuration $C$.

    When the metric is used in the assembly process, the two raw RGB images from the sensor outputs are evaluated immediately after the grip execution such that a decision on whether to proceed with the process in this grip configuration or if a regrip is needed can be made in real time.
    
    An encoder-decoder architecture, which is commonly used in the literature (see \cite{zeng2019tossingbot, Badrinarayanan2017}), is chosen to establish a causal relationship between $I$ and $\mathbf{w}_{\rm ext}^{max}$ so that $\bm{\xi}_i$ stays within $U_{r,\alpha}(\bm{\xi}_{i})$. 
    The encoder uses $I$ to extract the feature vector such that $f_{\rm Encoder}(I) \rightarrow \mathcal{F}$, and the decoder interprets these features to generate the prediction for $\mathbf{w}_{\rm ext}^{\rm max}$, i.e. $f_{\rm Decoder}(\mathcal{F}) \rightarrow \mathbf{w}_{\rm ext}^{\rm max}$. 
    As it is not apriori known as to which features $\mathcal{F}$ of $I$ would contribute to the stability of the grip, the encoder is needed to learn how potential features such as contact area, edges, or color gradients, are influencing the value of $\mathbf{w}_{\rm ext}^{\rm max}$.
	In comparison to the pixel-wise approaches where it is decided per pixel whether they contribute to a class (see \cite{Badrinarayanan2017}), the feature vector is translated into a vector quantity with this methodology.

	\section[EXPERIMENT FOR TRAINING DATA GENERATION]{Experimental Setup}\label{sec:experiment}
    To generate a reliable feature vector and to be able to interpret this vector, the NN needs a well-behaved and balanced dataset containing a variety of $I$ and associated maximal external 
	wrenches $\mathbf{w}_{\rm ext}^{\rm max}$
	which makes a specialized data generation method irreplaceable.
    
	Due to the parallel gripper used, the grip configuration $C$ defined in Section \ref{sec:methodology} can be simplified to $c=(\bm{\xi}_{\rm grip},F_{\rm grip})$, where the grip pose $\bm{\xi}_{\rm grip}$ is the center point between the two gripper jaws of the parallel gripper and $F_{\rm grip}$ is the Force applied by each gripper finger towards the center point. 
    For this experiment, as a proof-of-concept study, we simplified the described methodology by applying the external disturbance $\mathbf{w}_{\rm ext}=[0,0,F_z,0,0,0]$ exclusively along the z-axis. The maximum process force $F_{z}^{\rm max}$ at which the threshold $U=\Delta z\in \mathbb{R}$ is exceeded is analyzed, which is the special one-dimensional version of the neighborhood $U_{r,\alpha}$ defined in Section \ref{sec:methodology}.

    The goal of this experiment is to apply static forces to a gripped object to measure when it begins to slip out of the grasp at $\bm{\xi}_{\rm grip}$, as laid out in Section \ref{sec:methodology}. To decrease the degrees of freedom and to make the experiment more repeatable, the objects are fixed in all directions and orientations such that all forces applied by the robot on to the object can be treated as negative $F_{z}$.
	
	\subsection{Robotic System Setup}\label{sec:robot_setup}
	\begin{figure*}[t!]
		\centering
		\begin{subfigure}[b]{0.32\textwidth}
			\includegraphics[width=\columnwidth,trim={0 0 0 0},clip]{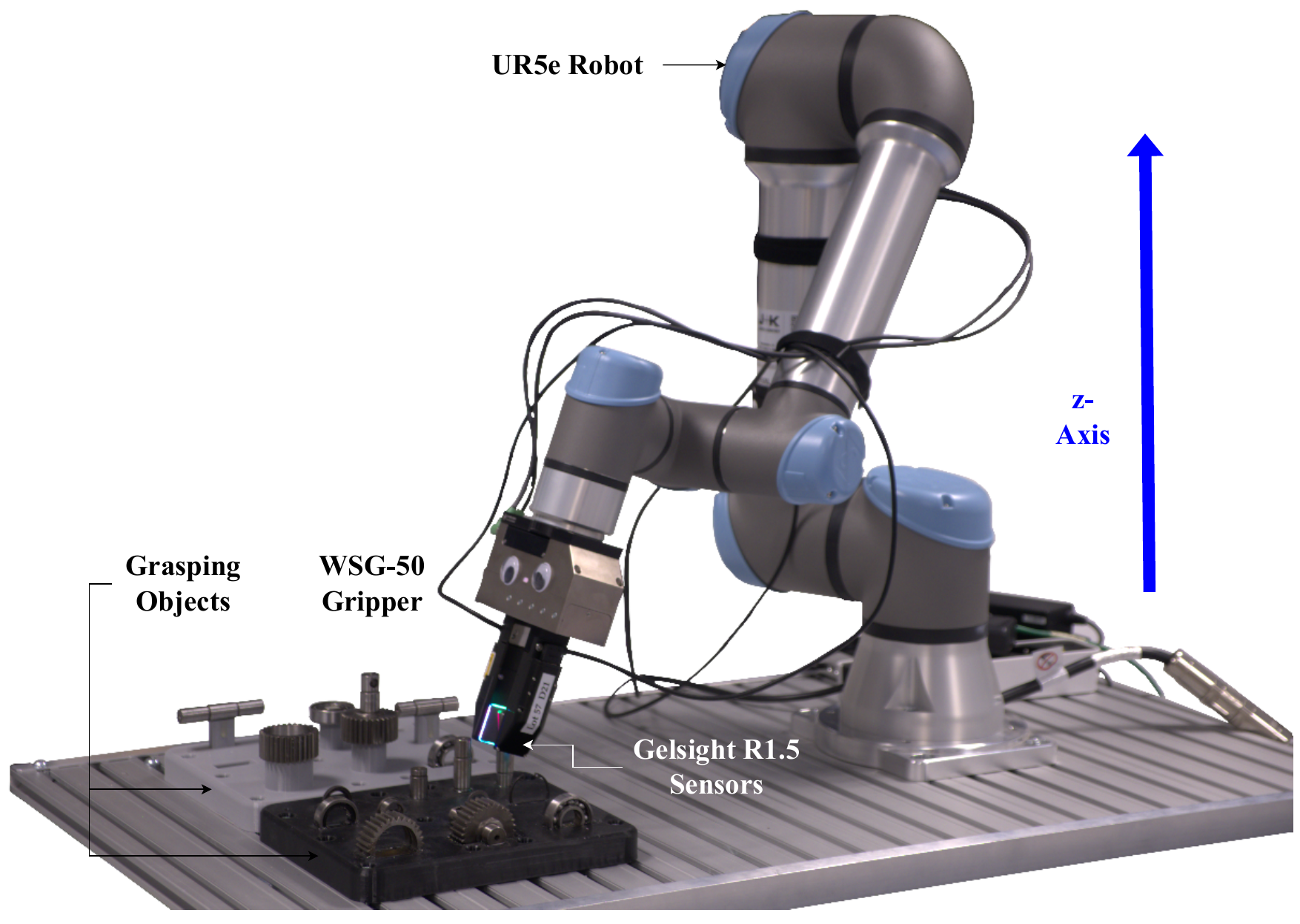}
			\caption{Experimental Setup}
			\label{fig:setup}
		\end{subfigure}
		\begin{subfigure}[b]{0.32\textwidth}
			\centering
			\includegraphics[width=\columnwidth]{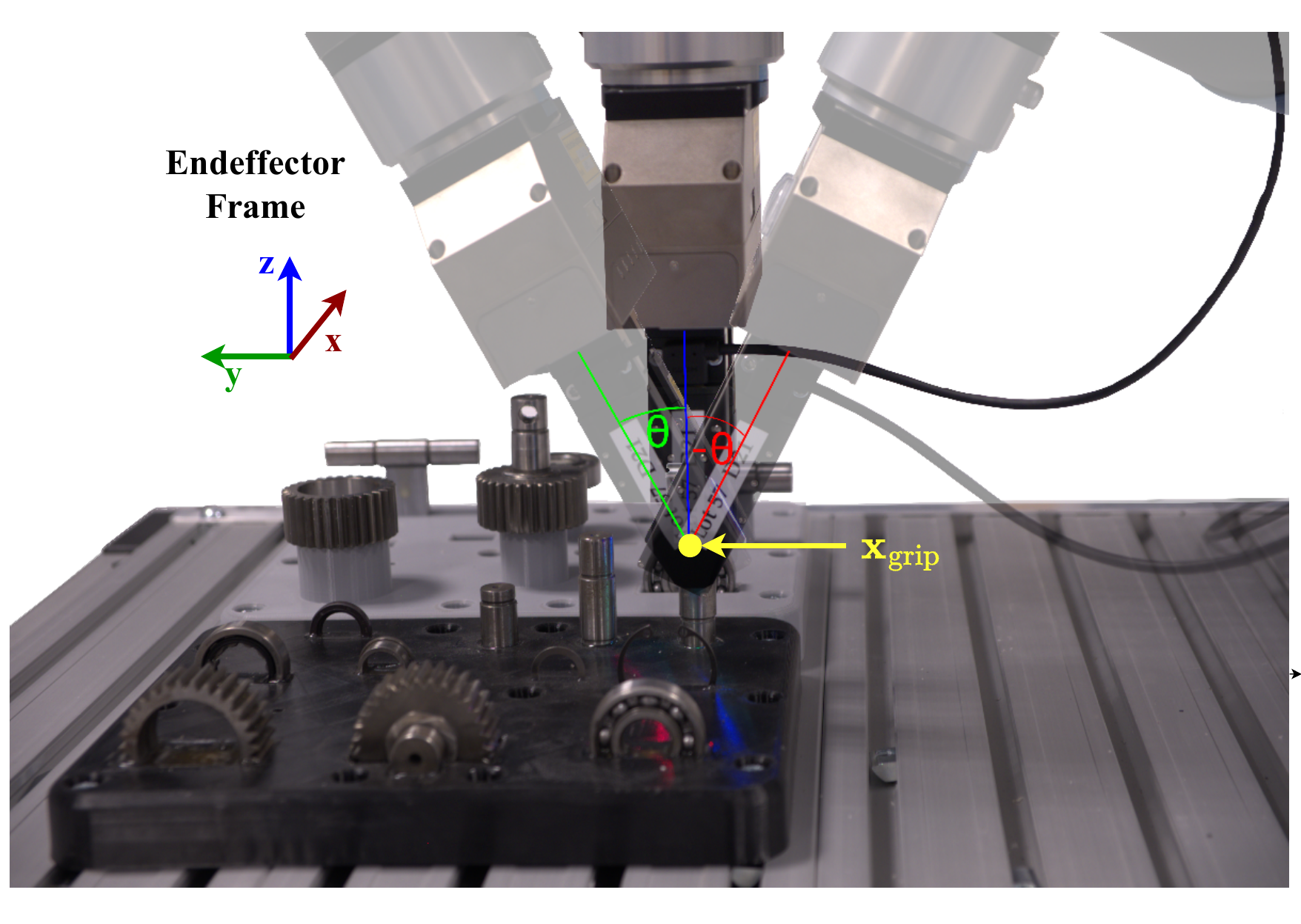}
			\caption{Angle Variance}
			\label{fig:anle-variance}
		\end{subfigure}
		\begin{subfigure}[b]{0.32\textwidth}
			\centering
			\includegraphics[width=\columnwidth,trim={0 0 0 0},clip]{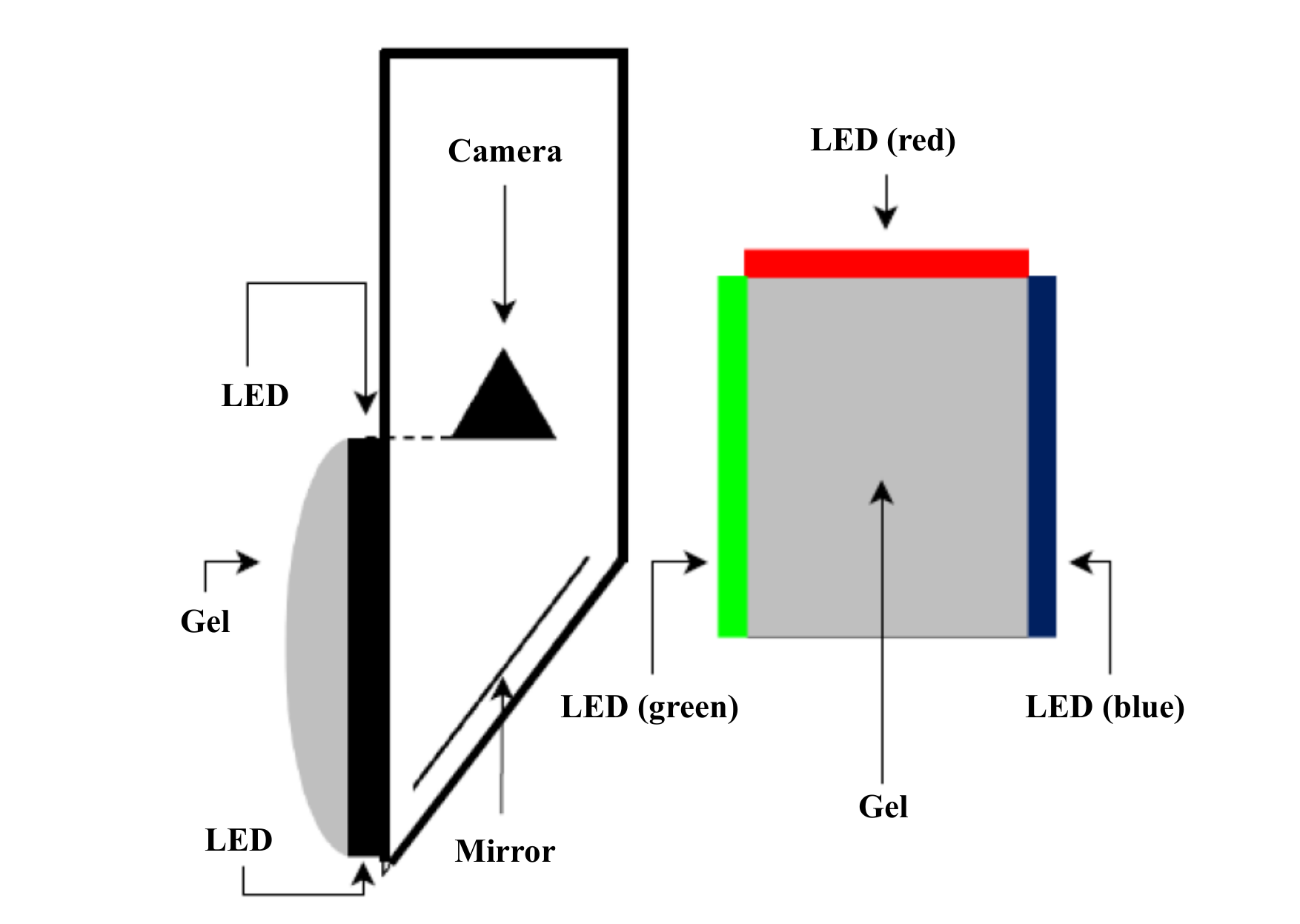}
			\caption{Basic Principle Gelsight}
			\label{fig:basic-principle-gelsight}
		\end{subfigure}
		\caption{The general robotic setup of the experiment.}
	\end{figure*}
	The mechanical measuring system consists of an UR5e robot, the WSG-50 parallel gripper, and two GelSight R1.5 optical tactile sensors. There is a recording board on the workbench of which the gearbox parts are fixed. This setup would be used to generate a varied reproducible industrial dataset (see Figure \ref{fig:setup}). 
    The UR5e is controlled using the Real-Time Data Exchange (RTDE), while the WSG-50 is controlled via its built-in telnet interface. The sensors measure $I$ by taking an image of the surface of a highly elastic gel. The gel is illuminated from three directions using three color channels. To make the sensor setup as compact as possible, the camera is located inside the R1.5 fingers, pointing towards the fingertip, where a mirror is aligned such that the gel image is reflected towards the camera. The image is then cropped, and the mirror effect is algorithmically removed, thus generating an RGB image with a resolution of $640$$\times$$480$ pixels. Figure \ref{fig:basic-principle-gelsight} shows a schematic view of the sensor setup.
	

	\subsection{Experimental Procedure}\label{sec:experimental_procedure}
	\begin{figure}[t!]
		\centering
		\includegraphics[width=\linewidth,trim={0 15cm 0 0},clip]{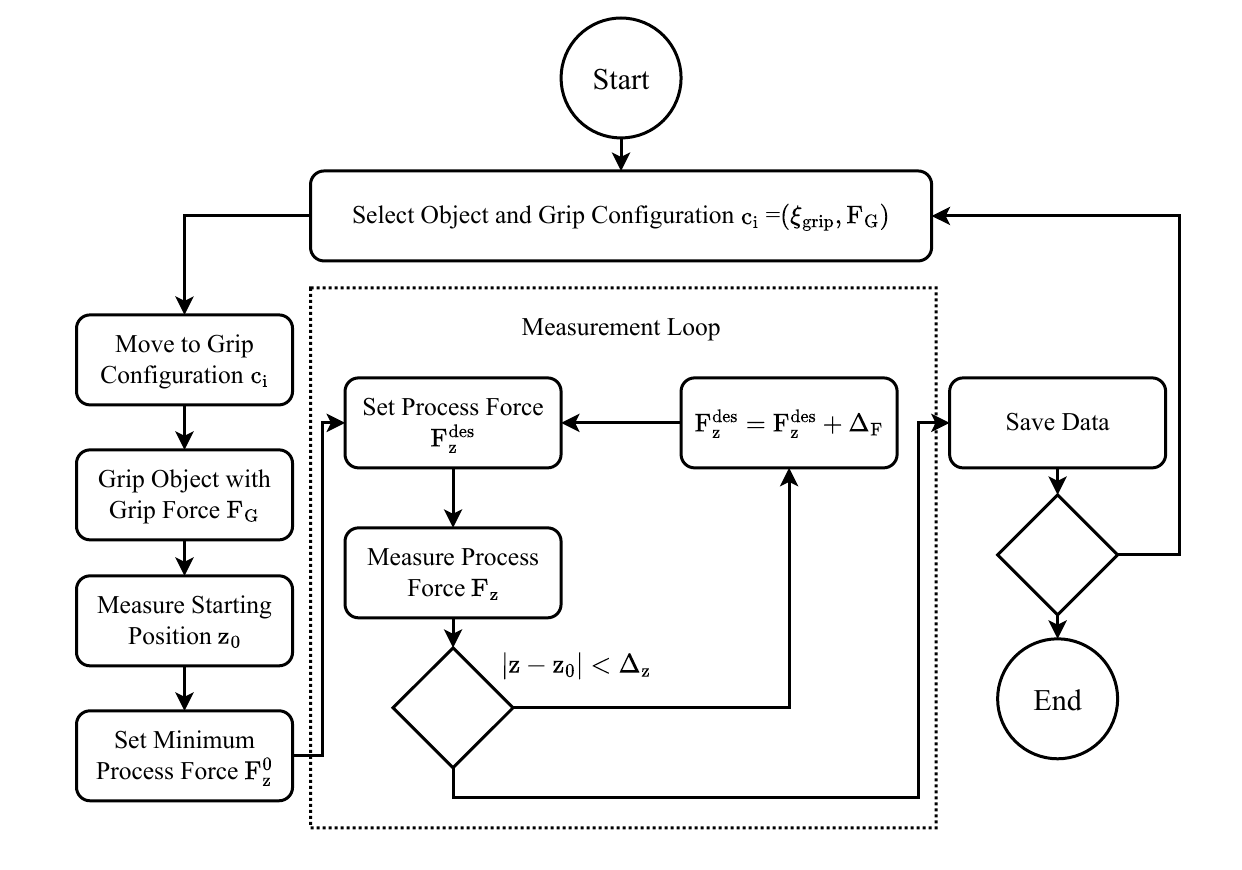}
		\caption{Flowchart for the data-taking procedure.}\label{fig:flowchart_experiment}
	\end{figure}
	\noindent The procedure of the experiment is as follows:
	\begin{enumerate}
		\item A robot grips a rigid object on the board with a specific grip configuration $c_i = (\bm{\xi}_{{\rm grip}_i},F_{G_i})$
		\item Grip images on both gripper jaws ($I_{\rm left},I_{\rm right})$, are recorded and stored.
		\item The robot starts to apply an external force along the z-axis with a step-like target pull force written using 
		\begin{equation*}
			F_{z}^{\rm des}(t) = \begin{array}{lr}
				F_{z}^0+i \Delta_F, & t \in [i\Delta_t, (i+1)\Delta_t), \end{array}
		\end{equation*}
		with constant force increments of $\Delta_F$ increases every $\Delta_t$~seconds and $i \in \mathbb{N}_0$.
		\item  The actual external force $F_{z}(t)$ is measured and logged together with $F_{z}^{des}(t)$. 
		\item As soon as the displacement of the robot exceeds the threshold $\Delta z$, the experiment is terminated.
	\end{enumerate}

    The variance of the valid grip poses is limited in the case of our experimental design and the WSG-50 parallel gripper used. In relation to the end effector frame, $\bm{\xi}_{\rm grip}$ can only be varied in the y-z plane with regard to the translation. Due to the one degree of freedom limitation of the WSG-50 gripper, i.e. the distance of between gripper jaws, the gripper cannot close if a variance along the x-axis is permitted. The orientation of the pose is similarly restricted. Complete freedom only exists with a rotation $\theta$ around the x-axis in the end effector system. An additional rotation around the y- and z-axes can only be carried out for very small angles. 
    
    To summarize, Steps 1--5 are repeated for different translations $(y_i, z_i)$ and rotations $\theta_i$ along the x-axis on different classes of objects and different grip forces $F_G \in [20,60]$~\SI{}{N}, with a step size of \SI{5}{N}, which results in the unique grip configuration $c_i=(\bm{\xi}_{{\rm grip}_i}, F_{{G}_i})$. Also, see Figure \ref{fig:flowchart_experiment} for the flowchart of the experimental procedure.

		\begin{table*}[t!]
			\caption{Overview of training and validation data.}\label{tab:train-val-data-overview}
			\centering 
			\begin{tabular}{ccp{2cm}p{2cm}p{2cm}c}
				\toprule
				\multicolumn{1}{c}{} & \textbf{Data Sample} & \multicolumn{3}{c}{\textbf{Objects}} & \textbf{Data Size ($|D_i|$)} \\ \midrule
				\multirow{2}{*}{\textbf{\begin{tabular}[c]{@{}c@{}}\\ \\Training \\ Samples\end{tabular}}} &
				  \multirow{2}{*}{\begin{tabular}[c]{@{}c@{}} \\ \vspace{2mm} \\ $D_T$ \end{tabular}} &
				  \centering Gear &
				  \multicolumn{1}{c}{Ball Bearing} &
				  \multicolumn{1}{c}{Axle Long} &
				  \multirow{2}{*}{\begin{tabular}[c]{@{}c@{}} \\ \vspace{2mm} \\ $2.0 \times 10^4$ \end{tabular}} \\
									 &                      & \centering \includegraphics[width=0.08\textwidth]{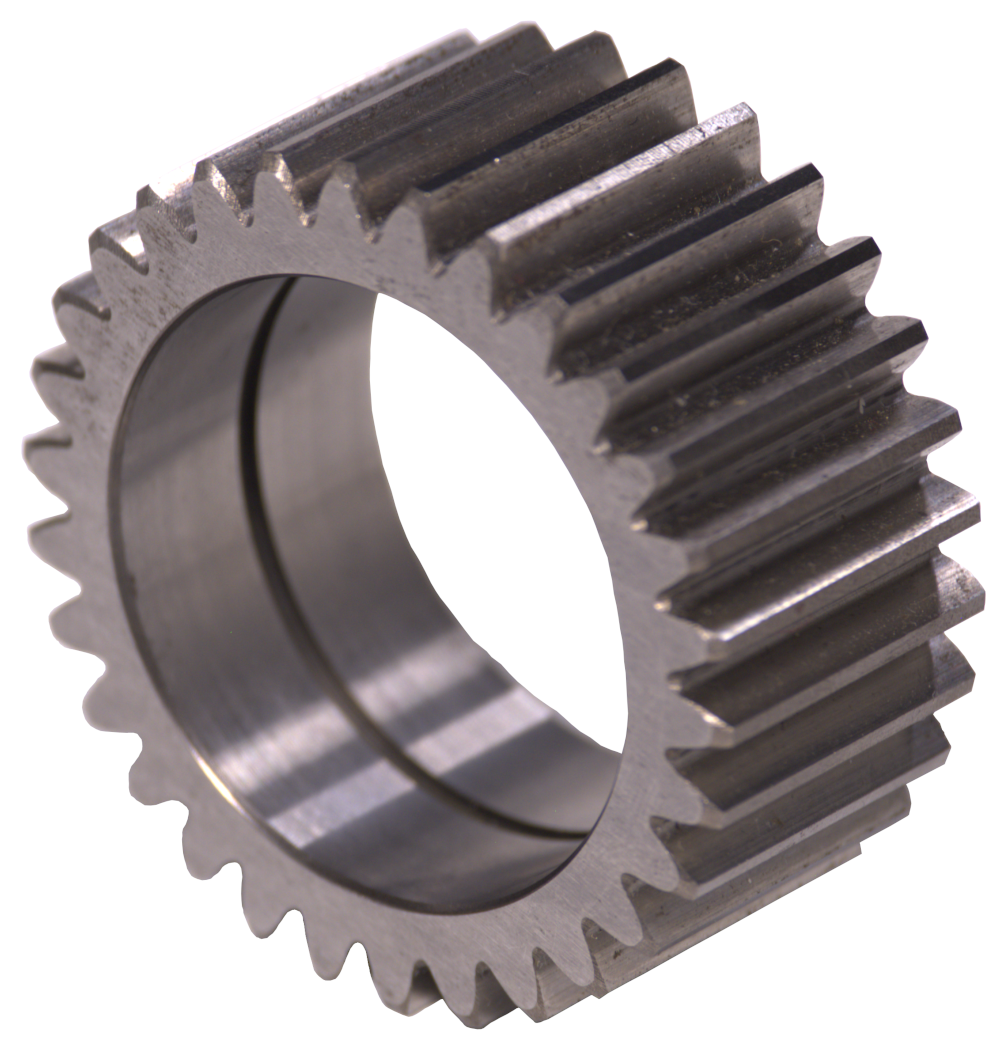}  & \centering \includegraphics[width=0.09\textwidth]{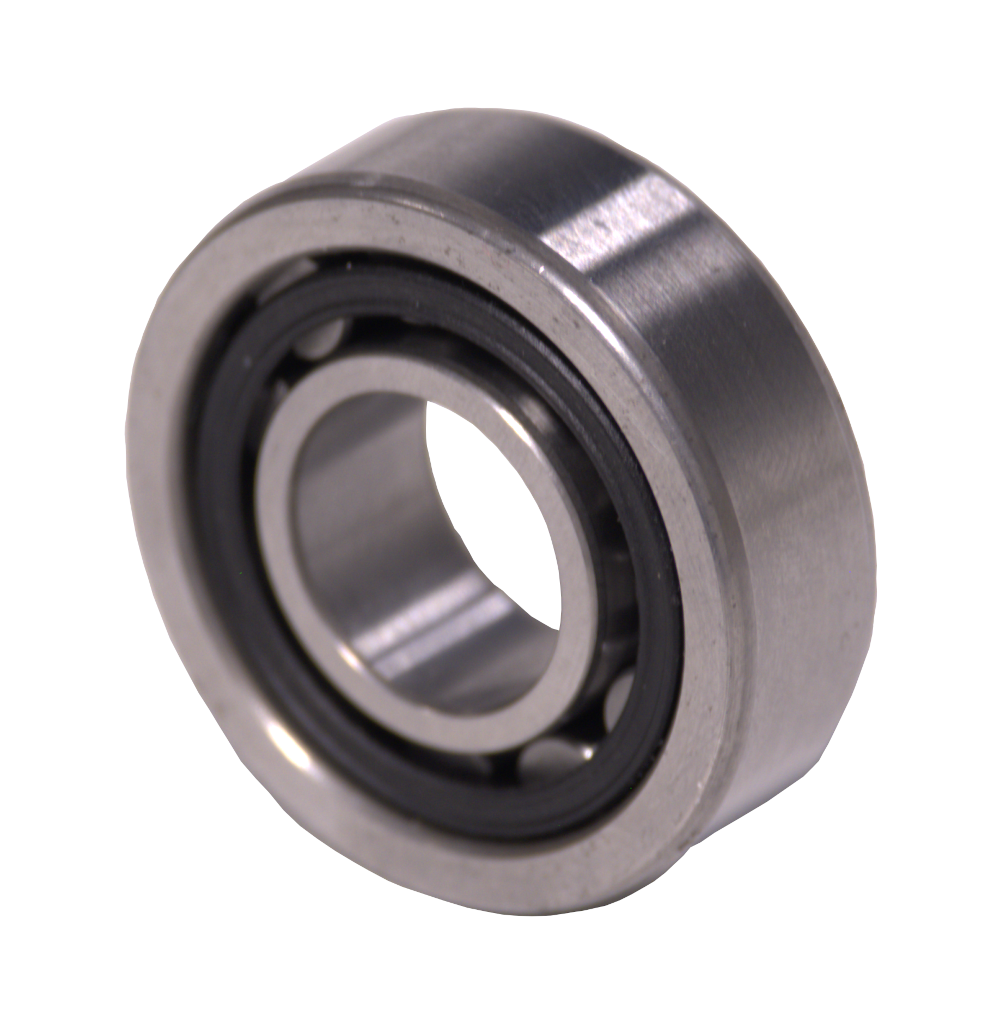}   & \centering \includegraphics[width=0.08\textwidth]{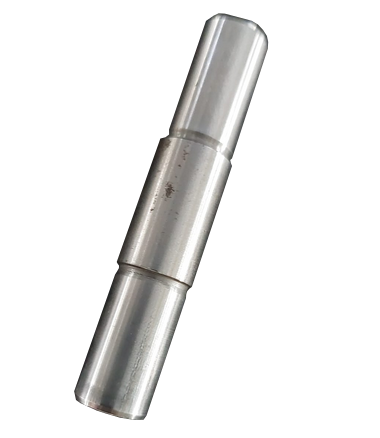}   &                              \\ \midrule
				\multirow{2}{*}{\textbf{\begin{tabular}[c]{@{}c@{}}\\ \\Validation \\ Samples\end{tabular}}} &
				  \multirow{2}{*}{\begin{tabular}[c]{@{}c@{}} \\ \vspace{2mm} \\ $D_V$ \end{tabular}} &
				  \centering Gear 1 &
				  \multicolumn{1}{c}{Gear 2} &
				  \multicolumn{1}{c}{Pinion Shaft} &
				  \multirow{2}{*}{\begin{tabular}[c]{@{}c@{}} \\ \vspace{2mm} \\ $2.0 \times 10^4$ \end{tabular}} \\
									 &                      & \centering \includegraphics[width=0.075\textwidth]{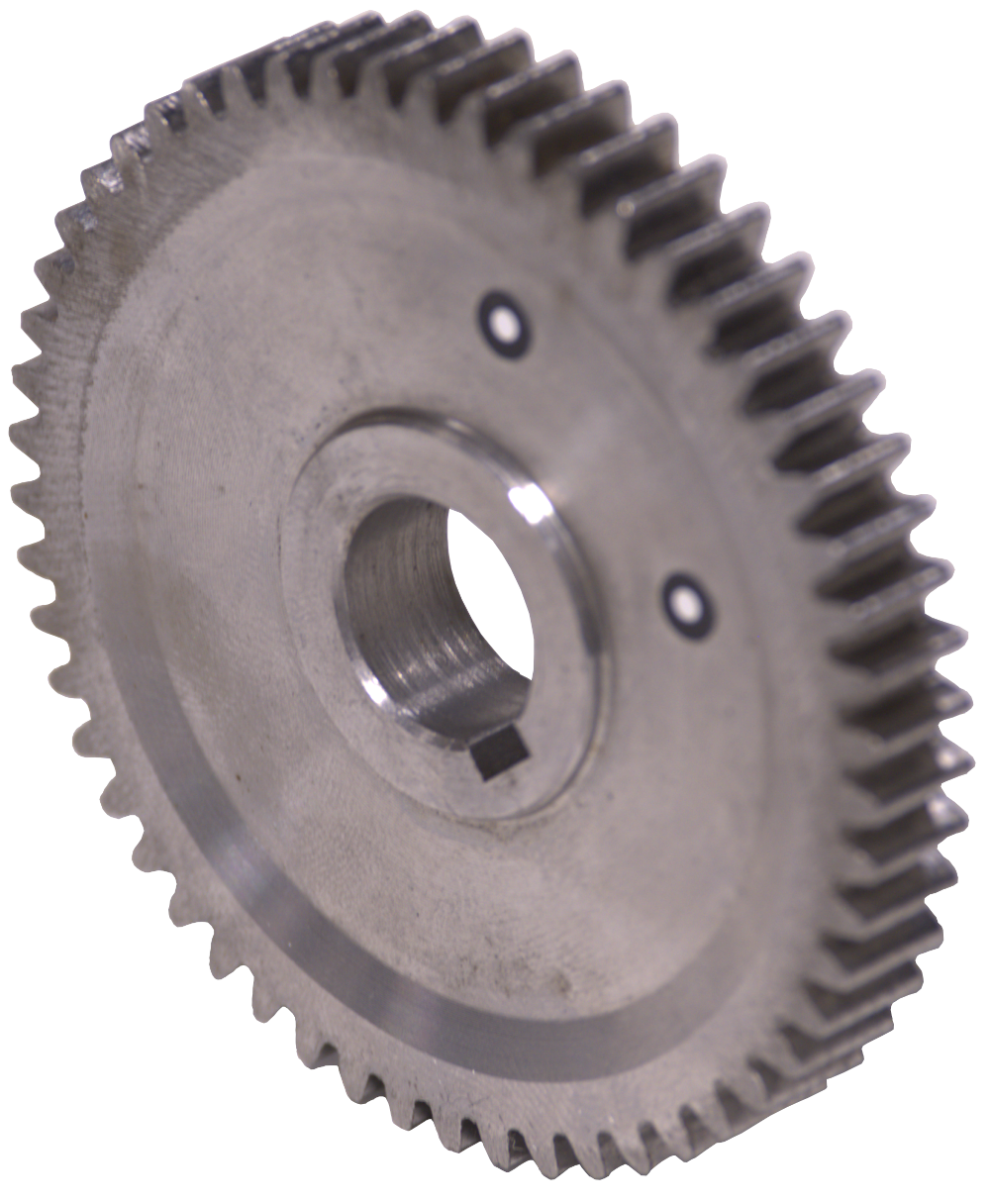}   & \centering \includegraphics[width=0.08\textwidth]{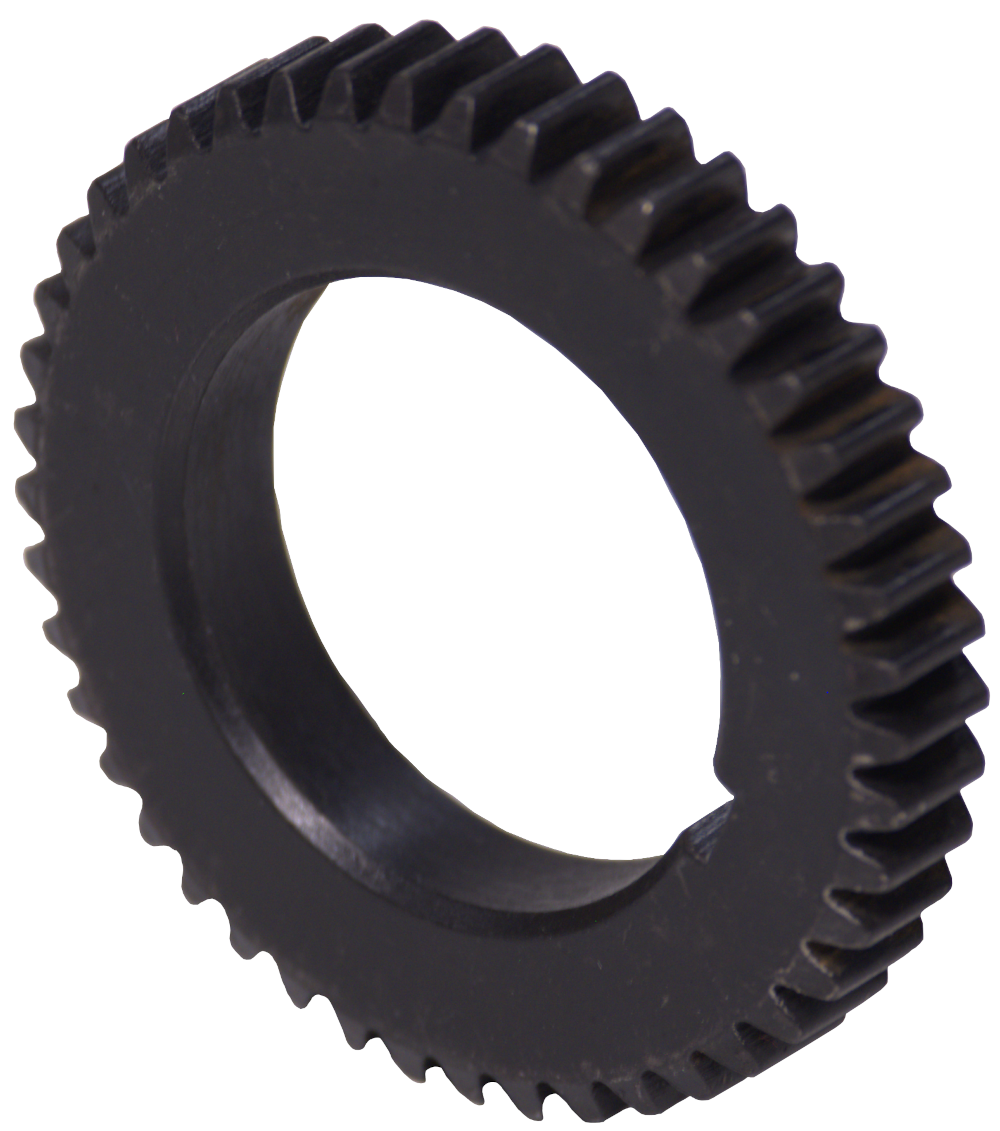}   &  \centering \includegraphics[width=0.12\textwidth]{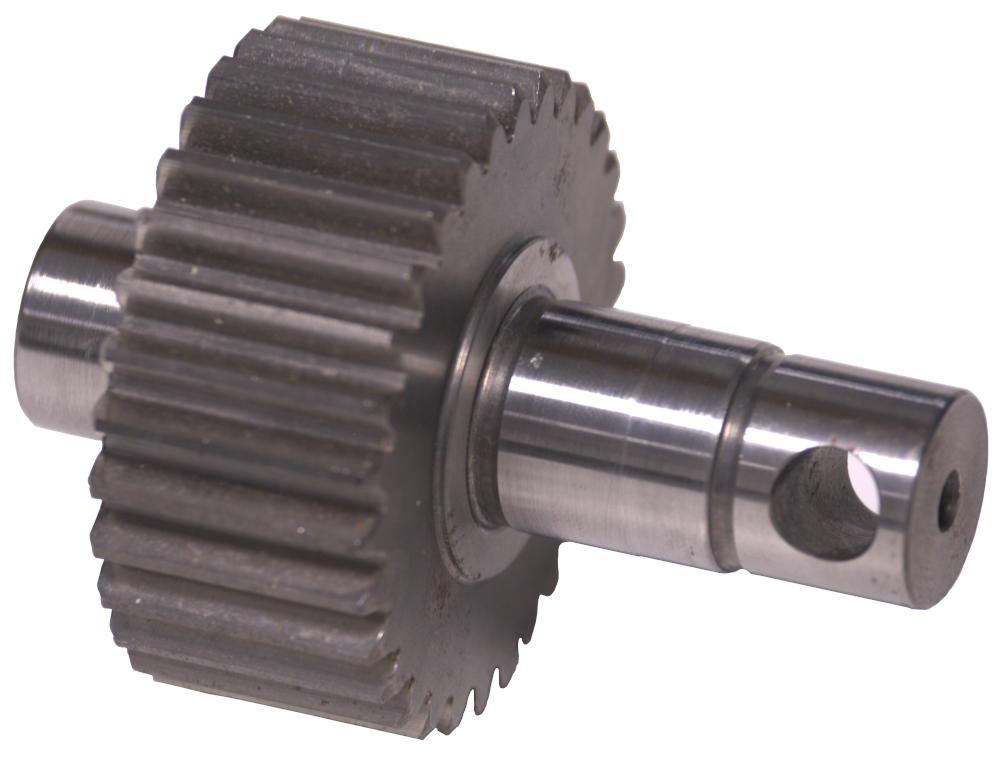}   &        \\ \bottomrule                     
			\end{tabular}
		\end{table*} 

	\subsection[Coupling of tactile info and stability]{Coupling Tactile Measurements and Pull Forces}\label{sec:label_calc}
	The aim of the experiment is to correlate the sensor images $(I_{\rm left},I_{\rm right})$ of the contact surface with $F_{z}^{\rm max}$ that can be applied to the object without slipping. 
	The process force $F_{z}^{\rm max}$ is scalar rather than vector like, as the wrench $\mathbf{w}_{\rm ext}^{\rm max}$ introduced in Section \ref{sec:experiment} is only nonzero in the Cartesian z-direction.
	The resulting metric should produce reliable predictions for rigid  objects made from different materials. 
	\begin{figure}[t!]
		\centering
		\includegraphics[width=0.9\linewidth]{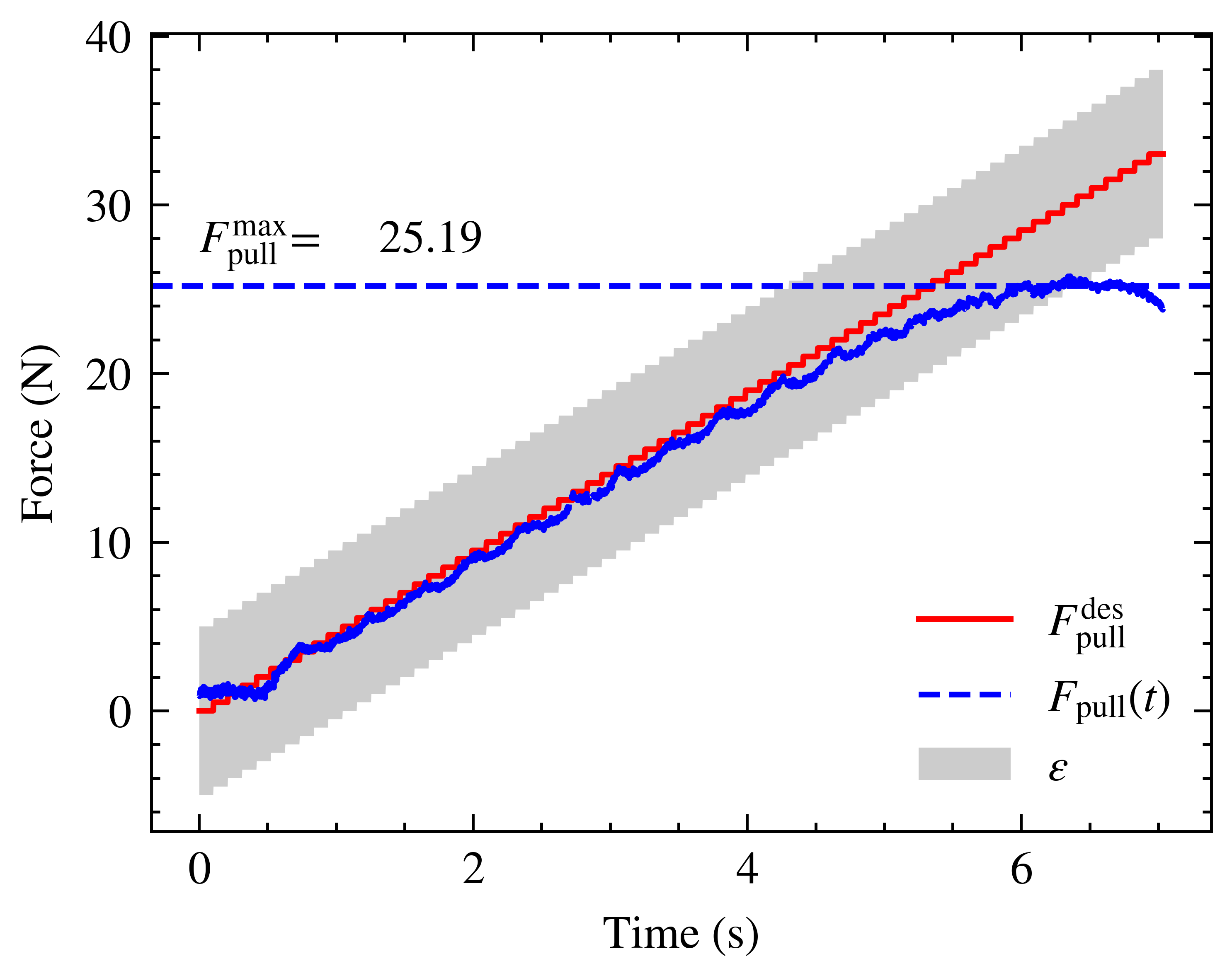}
		\caption{Determination of the maximum pull force.}\label{fig:max-pullforce-determination}
	\end{figure}

	Firstly, the experiment performed in Section \ref{sec:experimental_procedure} is used to determine $F_{z}^{\rm max}$. Figure \ref{fig:max-pullforce-determination} shows the plot of a typical experimental configuration, where $F_{z}^{\rm des}(t)$ is compared with $F_{z}(t)$ of the force-torque sensor in the last joint of the robot in the z-direction. Both forces are offset as the robot drivers have a reaction time before they can reach the set force. 
	Using the threshold $\varepsilon = \SI{3}{N}$, we can compute $F_{z}^{\rm max}$ using
	\begin{eqnarray}
		T_{\rm slip} &=& \arg \max_t\left\{\left|{F}_{z}(t)-F_{z}^{\rm des}(t)\right|<\varepsilon\right\},  \\
		F_{z}^{max}&=&F_{z}(T_{\rm slip}),
	\end{eqnarray}
	where $T_{\rm slip}$ is the slip time. 
	
	\subsection[Normalization of the Maximum Pull Force]{Normalization of the Maximum Pull Force}\label{sec:label_norm}
	Since NNs converge faster when datasets are normalized \cite{sola1997}, the maximum lateral force is regularized with respect to the upper $F_{max}$ and lower $F_{\rm min}$ force limits such that 
	\begin{equation}
		l_i = \frac{F_{z}^{\rm max}-F_{\rm min}}{F_{\rm max}-F_{\rm min}} \in [0,1].
	\end{equation} 
	In this paper, we set $F_{\rm max}=\SI{35}{N}$ and $F_{\rm min}=\SI{0}{N}$. 
	
	\subsection[Training and Validation Sample]{Training and Validation Sample}\label{sec:train_val_sample}
	
	The entire data sample $D$ consists of eight gearbox parts, such as different gears, ball bearings, pinion shafts, and axles. Due to their symmetries, points of contacts are selected for each part on the face and lateral surfaces. For every object and reference point, as described in Section \ref{sec:experimental_procedure}, a unique data point ${d_i} \in D$ is obtained for a specific grip configuration ${c_i}$. Each data point (consisting of an RGB image of the left and right GelSight tactile sensors and the corresponding normalized label) can be determined using
	\begin{equation}
		d_i=(I_{{{\rm RGB}_{\rm left}}_i},I_{{{\rm RGB}_{\rm right}}_i},l_{i}).
	\end{equation}

	There exist different datasets for training ($D_T$) and generalization ($D_V$) as disjoint subsets of the entire sample, where $D = D_T + D_V = 4.0\times10^4$.
	Table \ref{tab:train-val-data-overview} provides an overview of the different datasets, their sizes, and the objects considered.
	
	
	\section{Neural Network Design and Training}\label{sec:network_design}
	As discussed in Section \ref{sec:methodology}, we chose an encoder-decoder architecture for the design of a NN to predict the normalized stability $p_{i} \in [0,1]$ of $c=(\bm{\xi}_{\rm grip}, F_{G})$ on an unknown object. In our approach, the label $l_i$ for $c$ is a continuous measure, hence we choose a regression model. 
	This section presents our own architecture and a comparison model. 
	
	\subsection[Own Network Design - Stability Network]{Stability Neural Network (SNN) Design}\label{sec:own-network}
	Since wide ResNet architectures have prevailed over conventional CNNs in recent years \cite{Mascarenhas}, we decided to use ResNet blocks \cite{he2016deep} with a high number of filters within our model. 
	To identify the architecture and its parameters, hyperparameter tuning using hyperband was performed \cite{yu2020hyperparameter}. 
	
	
	\subsubsection[ResNet Block]{ResNet Block}\label{sec:resnet-block}
	Each ResNet block used consists of two convolutional layers with the same number of filters. Between the convolutional layers, a ReLu activation and batch normalization are performed. The number of filters in the block depends on the position of the block. After the convolutional layers, the input is added onto the output, using 0-padding on the original input if needed. Then, another set of ReLu activation and batch normalization are executed to generate the final output of the block.
	
	\subsubsection[Encoder]{Encoder}\label{sec:encoder}
	The encoder consists of a normal convolutional layer at the beginning, followed by a max pooling layer with stride 2, another ReLu, and batch norm. After this three ResNet blocks are; starting with 128 filters and doubling each time, with max pooling, ReLu, and batch norm in between the blocks. The output of the last block with 512 filters constitutes the output of the encoder. Each of the two RGB-images is passed to the encoder independently and the outputs of the encoder for both pictures are then concatenated in the filter dimension and passed to the decoder. 
	
	\subsubsection[Decoder-Feature Extraction]{Decoder - Feature Extraction}\label{sec:decoder-features}
	The decoder/interpreter receives the concatenated outputs of the encoder for both images, which is then passed through four additional ResNet layers, with the initial number of filters of 256. Batch normalization, max pooling, and ReLu activation are performed and the number of filters architecture halved block by block. 

	\subsubsection[Decoder-Interpreter]{Decoder - Interpreter}\label{sec:decoder-interpreter}
	The result after the ResNet layers is transformed into a vector and then transferred to four ReLu activated dense layers. The initial number of nodes is 512, which are halved after each layer. After the 2nd and 4th dense layers, dropout layers with a dropout rate of $0.5$ are part of the network. Finally, the output of the dense layer is fed to a layer with one node, which is then sigmoid activated to generate a normalized output.    
	
	\subsubsection[Training Parameters]{Training Parameters}\label{sec:training-params}
	The model is trained with the SAM optimizer \cite{foret2021sharpnessaware}, with $\beta = 0.05$, momentum $\rho = 0.9$ \cite{haji2021comparison}, learning rate $\eta = 0.1$, and batch size of 16. For our regression model, we use the MSE $\mathcal{L}_{\rm MSE}$ as the loss. 
	
	\subsection[Baseline Model]{Baseline Model}\label{sec:baseline-model}
	We use a ResNet-18 architecture as a comparison model. To obtain a 1-D scalar value for $p_i$, the output of the ResNet-18 model is converted into a vector and then the identical Decoder-Interpreter described in Section \ref{sec:decoder-interpreter} is applied. 
	The optimizer, loss function, accuracy metric, and other parameters are selected as discussed in Section \ref{sec:training-params}.

	\subsection[Training]{Training}\label{sec:training}
\begin{figure}[t!]
		\centering
		\begin{subfigure}[b]{0.9\columnwidth}
			\includegraphics[width=\columnwidth,trim={0 0 0 0},clip]{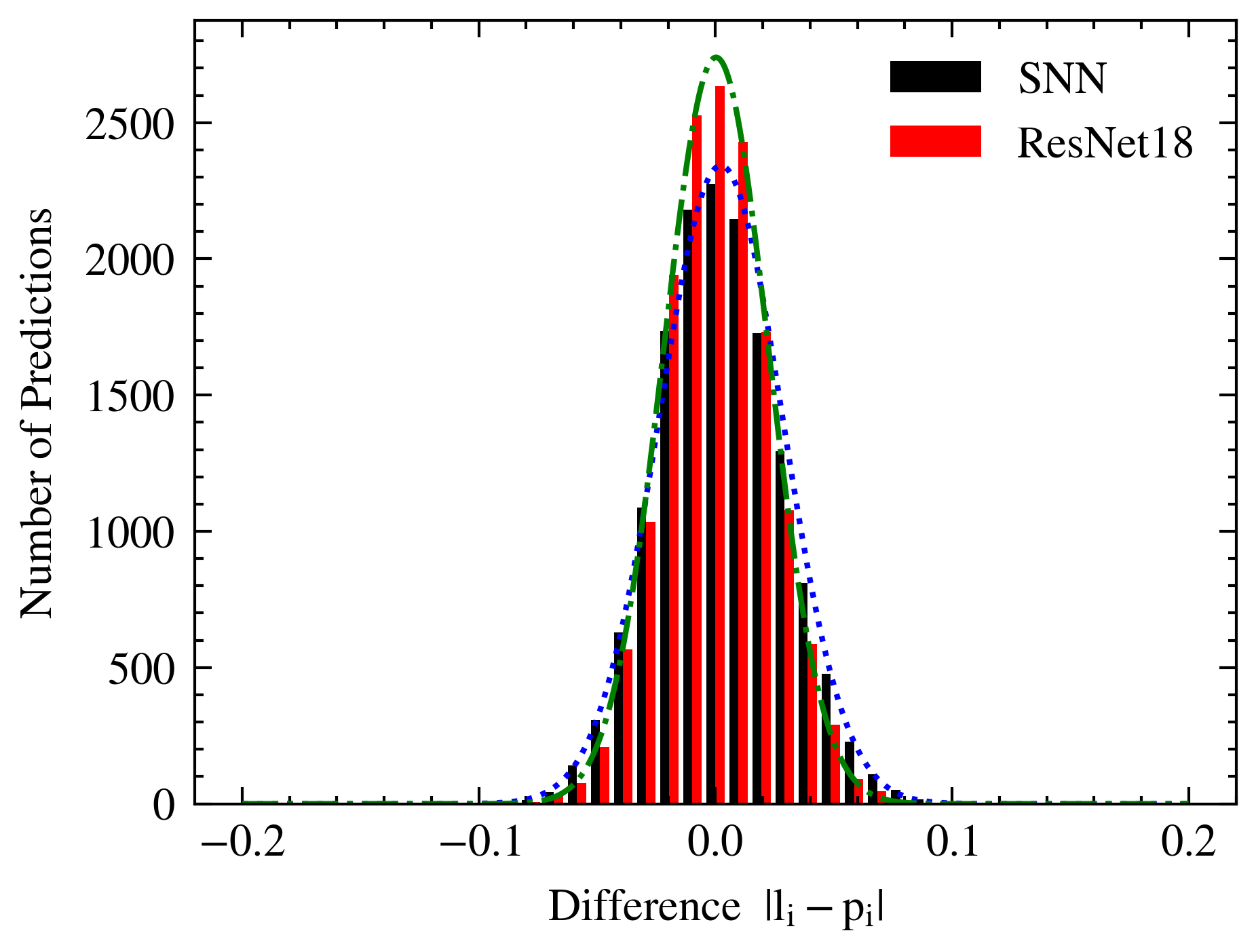}
			\caption{The distribution and Gaussian fit of the deviation between $l_i$ and $p_i$.}
			\label{fig:difference-nomalized}
		\end{subfigure}
		\begin{subfigure}[b]{0.9\columnwidth}
			\centering
			\includegraphics[width=\columnwidth]{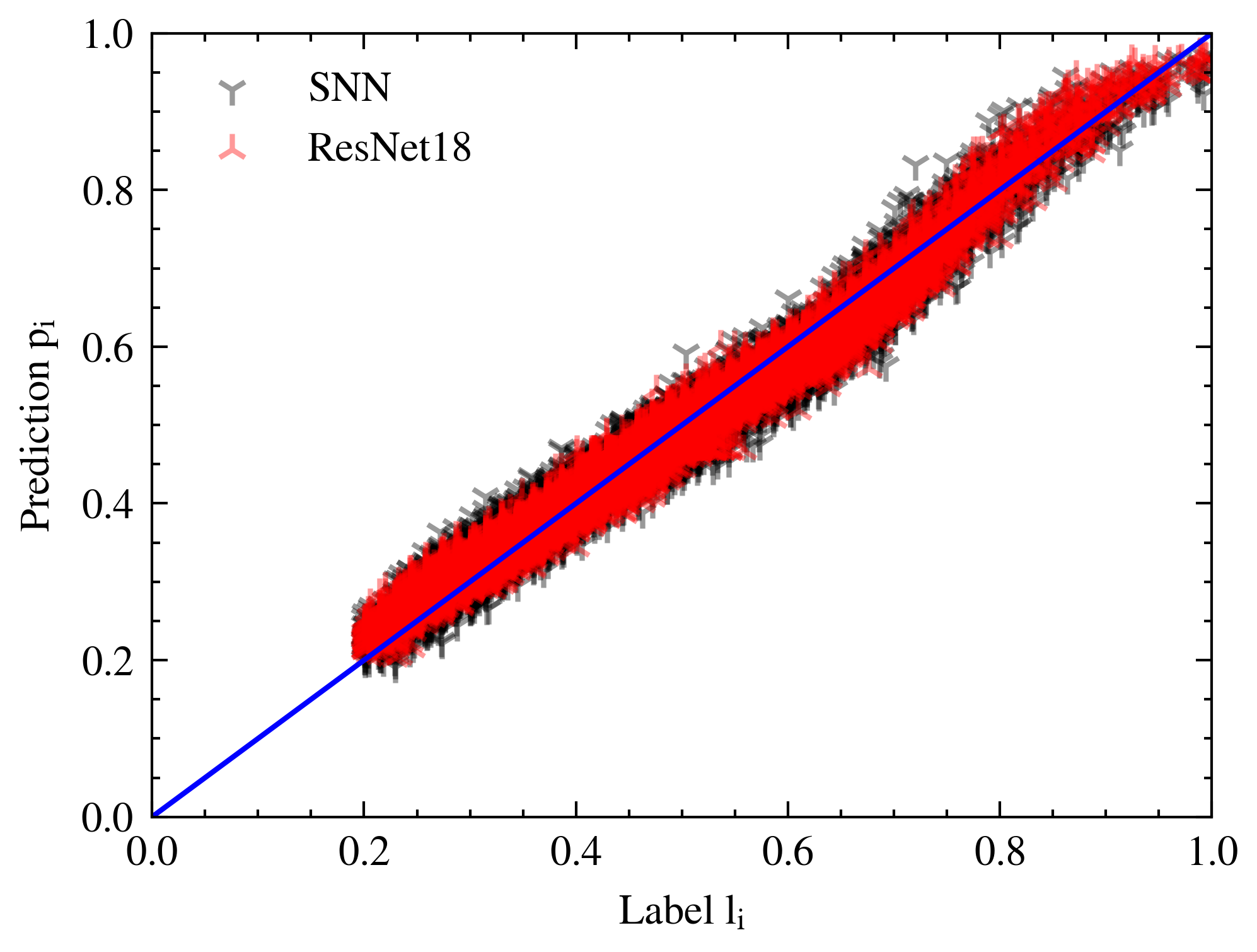}
			\caption{The scatter plot of the training results.}
			\label{fig:prediction-nomalized}
		\end{subfigure}
		\caption{Training results for the models using the dataset $D_T$.} 
		\label{fig:training-results}
	\end{figure}

	Both the stability network and the ResNet-18 model were trained from scratch. 
	To avoid overfitting and also to verify the representativeness and validity of the dataset, all models were trained using cross-validation. 
	Based on the size of our training datasets, a $1:10$ ratio of K-folds was formed as suggested in \cite{snee1977,afendras2019optimality}. 
	For training, the dataset $D_T$ described in Section \ref{sec:train_val_sample} was used.   
    To assess the quality of the models, we use the accuracy $\mathcal{A}_{\rm mean}$ and precision $\mathcal{P}_{\rm RMSE}$ of the deviations between label $l_i$ and prediction $p_i$. They are defined as
    \begin{eqnarray}
		\mathcal{A}_{\rm mean} &=& \frac{1}{N}\sum_{i}^{N} (l_{i}-p_{i}), \\
		\mathcal{P}_{\rm RMSE} &=& \sqrt{\frac{1}{N-1}\sum_{i}^{N} ((l_{i}-p_{i})-\mathcal{A}_{\rm mean})^2}.
	\end{eqnarray}
    Since both the accuracy and the precision are normalized, we present them as forces with the unit N by multiplying them with the normalization factor $F_{\rm max}$ defined in Section \ref{sec:label_calc} for easier interpretation of the results.
    The results in the Tables \ref{tab:trainresults}, \ref{tab:generalization-results} are therefore presented as:
	\begin{equation}
		F_{{\mathcal{A}}}\pm F_{{\mathcal{P}}} = ({\mathcal{A}_{\rm mean}} \pm \mathcal{P}_{RMSE}) 
        F_{\rm max},
	\end{equation} 
	with ${F_{\mathcal{A}}}$ and ${F_{\mathcal{P}}}$ the accuracy and precision described as forces. 
    The results are of high accuracy and precision if ${F_{\mathcal{A}}}$ and ${F_{\mathcal{A}}}$ are close to zero, since label $l_i$ and prediction $p_i$ are close to each other, (see the histogram in Figure \ref{fig:difference-nomalized}). 
	Table \ref{tab:trainresults} and Figure \ref{fig:training-results} presents the results of the trainings. SN represents Stability Network, and ResNet-18 is the comparison model. 
	The mean values of the training results show a maximum deviation of \SI{0.1}{N} and the maximum fluctuation of the data is \SI{1.2}{N}. The noise of the force sensor in the last joint of the UR5e robot is $\pm$\SI{4}{N}, hence all models are of high accuracy. Figure \ref{fig:difference-nomalized} shows that the distributions are Gaussian, indicating that the experimental procedure yields a well-behaved dataset for training, especially since no outliers can be identified.
	\begin{table}[t!]
		\caption{Training results using the $D_T$ dataset presented as forces as described in Section \ref{sec:training}. SNN represents the Stability Neural Network. 
        The results are presented as forces as described in Section \ref{sec:training}.}
		\label{tab:trainresults}
		\centering 
		\begin{tabular}{@{}cc@{}}
			\toprule
			\textbf{Model}       & \textbf{Training Results $[$\SI{}{N}$]$}  \\ \midrule
			\textbf{SNN}          & $0.077 \pm  0.91$          \\
			\textbf{ResNet-18}   & $\mathbf{0.0095 \pm 0.81}$ \\ \bottomrule
			\end{tabular}
	\end{table}

	\section{Validation}\label{sec:validation}
	The trained models were evaluated based on their generalization potential. In concrete terms, the models were analyzed for their robustness of unknown features in the tactile grip images using the validation dataset $D_V$. This is illustrated in Table \ref{tab:train-val-data-overview}, where the dataset consists of unknown gears and a pinion shaft. 
	A distinction was made between models that only received the grip images and those also using $\theta$ as input.  
	
	
	Table \ref{tab:generalization-results} shows the results of the generalization test with the best performing models highlighted in bold. 
	Both networks performed similarly, with an accuracy of less than \SI{1}{N} (except for SNN with up to \SI{2.1}{N}) off the label value and a prediction precision of less than \SI{3.4}{N}. This spread is smaller than the \SI{4}{N} measurement uncertainty of the force sensor on the robot. The results vary depending on the different objects, with the networks generally performing worse on Gear~1 than on Gear~2 or the pinion shaft. These results provide some insights into how the networks find similarities between objects. The pinion shaft is comparable to an axle, as both are cylindrical. Gear~2 contains a roughly similar ring structure as the gears in $D_T$, albeit with a different diameter. Finally, the images of Gear~1 contain additional unknown structures that might influence the prediction away from the actual value.
	
	Nevertheless, the networks generalize well to the objects that contain features encountered previously (i.e., Gear~2 and pinion shaft), while it generalizes more weakly, but still performs better than the measuring uncertainty, for objects with unknown structures (i.e., Gear~1). 
	
	\begin{table}[t!]
		\caption{Comparison of the models on their generalization potential on $D_V$. SNN represents the Stability Network 
        The results are presented as forces as described in Section \ref{sec:training}.}\label{tab:generalization-results}
		\centering 
		\begin{tabular}{cccc}
			\toprule
			\multirow{2}{*}{\textbf{\textbf{Model}}} & \multicolumn{3}{c}{\textbf{Validation Results $[$\SI{}{N}}$]$}  \\ \cline{2-4} 
						  & \textbf{Gear 1} & \textbf{Gear 2} & \textbf{Pinion Shaft}  \\ \midrule
			\textbf{SNN}   & $2.1 \pm 3.4$            & $1.2\pm 1.8$    & $-1.1\pm 1.9$          \\
			\textbf{ResNet-18}                       & $\mathbf{-0.56 \pm 2.9}$ & $\mathbf{0.043\pm 1.8}$ & $\mathbf{-0.35\pm 1.65}$ \\ \bottomrule
			\end{tabular}
	\end{table}
	
	
	\section{Conclusion}\label{sec:conclusion}
	This paper presented a new procedure to automatically create and label a dataset for training models to estimate grip stability using GelSight sensors. Compared to other methods, the dataset creation method uses force measurements to generate a continuous label. In addition, a new NN structure is also introduced to transform two input images into force values using an encoder-decoder structure. This structure was compared with a ResNet-18 model. It was found that both architectures produce predictions compatible with each other in their respective precision.
	Both models generalize well to objects that exhibit the same general features as the objects used for training, but the prediction becomes worse if new features are encountered. Importantly, the predicted force varies by less than the measurement accuracy of the force torque sensors of the UR5e robot.
	
	Potential future work includes generalizing the experimental setup for more complex wrenches. 
	Additionally, the data generation procedure should be improved to reduce wear on the sensor hardware and speed up acquisition times via simulation, which would require further research in the simulation of both contact dynamics and the response of the GelSight sensors to be carried out.  
	
	\bibliographystyle{unsrt}
	\bibliography{bibliography}
\end{document}